\documentclass{article}

\PassOptionsToPackage{numbers}{natbib}
\usepackage[final]{nips_2017}

\usepackage{fancyvrb}
\usepackage{xcolor}

\usepackage[utf8]{inputenc} 
\usepackage[T1]{fontenc}    
\usepackage{hyperref}       
\usepackage{url}            
\usepackage{booktabs}       
\usepackage{amsfonts}       
\usepackage{nicefrac}       
\usepackage{microtype}      
\usepackage{graphicx}
\usepackage{acronym}
\usepackage{xspace}
\usepackage{amsmath,amssymb,amsfonts}
\usepackage{algorithmic}
\usepackage{multirow}
\usepackage{makecell}
\usepackage{graphicx}
\usepackage{tablefootnote}
\usepackage{subcaption}
\newcommand{\eg}{e.g.,\xspace}

\newcommand{\etal}{et\ al.\xspace}

\title{BiSparse-AAS: Bilinear Sparse Attention and Adaptive Spans Framework for Scalable and Efficient Text Summarization}

\author{
  Desta Haileselassie Hagos \\
  Howard University, USA \\
  \texttt{desta.hagos@howard.edu} \\
  \And 
  Legand L. Burge \\
  Howard University, USA \\
  \texttt{lburge@howard.edu} \\
  \And
  Anietie Andy \\
  Howard University, USA \\
  \texttt{anietie.andy@howard.edu} \\
  \And
  Anis Yazidi\\
 University of Oslo, Norway\\
  \texttt{anisy@ifi.uio.no} \\
  \And
  Vladimir Vlassov\\
  KTH Royal Institute of Technology, Sweden\\
  \texttt{vladv@kth.se} \\
}

\acrodef{NLP}{Natural Language Processing}
\acrodef{RoPE}{Rotary Position Embedding}
\acrodef{LLMs}{Large Language Models}

\begin{document}

\maketitle

\begin{abstract}
  Transformer-based architectures have advanced text summarization, yet their quadratic complexity limits scalability on long documents. This paper introduces BiSparse-AAS (\textit{Bilinear Sparse Attention with Adaptive Spans}), a novel framework that combines sparse attention, adaptive spans, and bilinear attention to address these limitations. Sparse attention reduces computational costs by focusing on the most relevant parts of the input, while adaptive spans dynamically adjust the attention ranges. Bilinear attention complements both by modeling complex token interactions within this refined context. BiSparse-AAS consistently outperforms state-of-the-art baselines in both extractive and abstractive summarization tasks, achieving average ROUGE improvements of about 68.1\% on CNN/DailyMail and 52.6\% on XSum, while maintaining strong performance on OpenWebText and Gigaword datasets. By addressing efficiency, scalability, and long-sequence modeling, BiSparse-AAS provides a unified, practical solution for real-world text summarization applications.
\end{abstract}

\section{Introduction}
\label{introduction}

Automatic text summarization has become a critical task in \ac{NLP}, driven by the exponential growth of digital content and the need to extract concise and coherent summaries from large amounts of information. Text summarization techniques are broadly categorized into two types: \emph{abstractive} and \emph{extractive}~\cite{see2017get}. Abstractive summarization generates original summaries by synthesizing, rephrasing, and paraphrasing information from the source text, leveraging advanced \ac{NLP} techniques. Unlike extractive methods, abstractive summarization does not rely on replicating sentences directly from the source~\cite{see2017get}. These techniques leverage transformers and \ac{LLMs} to produce semantically meaningful and coherent text sequences~\cite{sutskever2014sequence, tan2017abstractive}. However, abstractive summarization is computationally more demanding than extractive methods because it relies on advanced generative capabilities and decoder-based architectures~\cite{see2017get, gehrmann2018bottom, lewis2020bart}. Extractive summarization, on the other hand, identifies and selects key sentences or phrases directly from the source text~\cite{see2017get, kupiec1995trainable}. By directly copying content, this approach ensures grammatical correctness and relevance but lacks the ability to rephrase or condense information effectively. Modern extractive techniques may leverage neural networks to identify key sentences, but avoid the complexities inherent to generative processes required by abstractive summarization.

Traditional summarization approaches, such as statistical and rule-based techniques, have largely been replaced by neural network models, particularly those based on the Transformer architecture. The introduction of Transformer-based models by Vaswani \etal~\cite{vaswani2017attention} revolutionized \ac{NLP} with the self-attention mechanism, enabling parallel sequence processing and achieving state-of-the-art performance. However, despite their success, Transformer models face significant challenges due to the quadratic complexity $O(n^2)$ of the self-attention mechanism with respect to the input sequence length $n$. This results in memory usage scaling quadratically, meaning that memory requirements quadruple as sequence lengths double. Such scaling makes handling sentences or documents computationally expensive and impractical in resource-constrained environments. Although models such as BART~\cite{lewis2020bart} and PEGASUS~\cite{zhang2020pegasus} have advanced pretraining and fine-tuning techniques, they do not fully resolve the computational inefficiencies of processing long inputs. These models often struggle to maintain context and coherence for longer sequences, and their high parameter counts require substantial computational resources, limiting their applicability in real-world scenarios with resource-constrained environments. Although parameter count and computational efficiency are related, they are not always strongly correlated. For example, a smaller Transformer with long input sequences still suffers from the quadratic complexity of the self-attention mechanism, whereas a larger Transformer with sparse attention can handle long sequences more efficiently. Thus, efficiency depends more on the sequence length, attention mechanisms, and architectural optimizations than on the parameter count alone. Addressing these challenges is crucial for developing efficient and scalable summarization systems while preserving the integrity of the input text.

Model performance often degrades for sequences beyond 1024 tokens due to the limitations in modeling long-range dependencies and maintaining contextual coherence. This limitation has prompted the development of various techniques aimed at extending the model's context window without sacrificing efficiency.  Techniques like RoPE~\cite{su2024roformer}, employed in modern \ac{LLMs}~\cite{liu2023scaling}, enhance positional encoding for longer sequences, addressing the limitations of fixed position encodings in earlier Transformer models. These advancements improve performance on tasks that require extended context windows. However, while RoPE scaling effectively extends the context length, it does not fully resolve the challenges associated with computational efficiency and parameter optimization. As a result, tasks involving long inputs continue to face challenges in scalability and efficiency.  

This ongoing gap highlights the need for comprehensive solutions that balance context extension, computational efficiency, and scalability in Transformer-based models. Addressing these challenges motivates the development of methods that not only improve long-sequence handling but also optimize performance in resource-constrained environments, enabling more practical and effective text summarization systems. Compared to previous work on summarization~\cite{zhang2020pegasus,liu2019text,lewis2020bart}, BiSparse-AAS addresses key limitations in efficiency (quadratic complexity), scalability (fixed attention spans), and context preservation (limited handling of long-range dependencies). These advances enable improved performance on long-document summarization tasks, as detailed in Section~\ref{Our_Proposed-Approaches}.

\section{Motivation}
\label{motivation}

The limitations of existing state-of-the-art Transformer-based models in text summarization tasks, particularly in handling long sequences, preserving context, and optimizing computational efficiency, highlight the need for innovative approaches to address these challenges. While advancements such as \ac{RoPE} scaling~\cite{liu2023scaling} have extended the context window of models, they do not fully address computational inefficiencies and context preservation challenges. Existing models, including Transformer~\cite{vaswani2017attention}, BART~\cite{lewis2020bart}, and PEGASUS~\cite{zhang2020pegasus}, demonstrate strong performance in many scenarios but struggle to summarize long documents due to their quadratic complexity. Recent models like Longformer~\cite{beltagy2020longformer}, BigBird~\cite{zaheer2020big}, and Longformer-Encoder-Decoder (LED)~\cite{beltagy2020longformer}, are specifically designed for efficient long-sequence handling using sparse attention mechanisms. However, although these methods are computationally efficient, they often face trade-offs in context preservation and model expressiveness, which are critical for accurate and meaningful summarization. Other efficient attention variants, such as Performer~\cite{choromanski2021rethinking}, approximate the self-attention mechanism with linear complexity through kernel-based methods rather than sparse attention, but still struggle to model long-range dependencies effectively.

To address these limitations, we propose BiSparse-AAS, a novel framework that integrates parameter-efficient bilinear attention, sparse attention, and adaptive attention spans. Bilinear attention optimizes the self-attention mechanism by reducing the parameter count while capturing complex token relationships. Sparse attention improves efficiency by focusing computational resources on the most relevant parts of the input, and adaptive spans dynamically adjust attention ranges to preserve critical contextual information in long documents. For this task, our approach reduces the parameter count from 124 million (GPT-2) to approximately 102 million, improving efficiency while maintaining strong performance. Together, these components enable efficient processing of long sequences, providing a scalable and context-aware solution for text summarization. Unlike existing methods such as \ac{RoPE} scaling, which primarily extends the context window, BiSparse-AAS ensures that this extended capacity is utilized effectively, enhancing both performance and scalability in real-world applications.

In this paper, we demonstrate the applicability of BiSparse-AAS to text summarization and propose a unified framework that is suitable for both extractive and abstractive models. The proposed framework is well-suited for real-world applications with constrained computational resources, enabling practical deployment across diverse scenarios. Our implementation adopts a base context window of 1024 tokens, which is consistent with the original GPT-2 model~\cite{karpathy2022nanogpt}. We aimed to develop a GPT-2-style model due to its lower resource demands, ease of deployment, and suitability for fine-tuning. In our opinion, GPT-2 provides a practical and cost-effective solution for this task. Although current experiments do not explicitly extend this window, the sparse attention and adaptive span mechanisms are designed to efficiently handle longer sequences, with the potential to extend the effective context window beyond 1024 tokens through further optimizations and hardware capabilities. BiSparse-AAS represents a significant advancement in text summarization by combining adaptability, computational efficiency, and contextual awareness to address the longstanding limitations of Transformer-based models.

Beyond text summarization, BiSparse-AAS provides a general framework for improving the computational efficiency and scalability of sequence modeling. By addressing the challenges of long-sequence handling and computational costs, it offers a structured solution that is applicable to both extractive and abstractive summarization tasks.

\vspace{1.0ex}
\noindent \textbf{Contributions}. Our paper makes the following contributions. 

\begin{itemize}

    \item We propose BiSparse-AAS, a novel framework that integrates bilinear attention, sparse attention, and adaptive spans, offering an efficient and scalable solution for language modeling and text summarization.

    \item We evaluate BiSparse-AAS on diverse datasets, including CNN/DailyMail~\cite{hermann2015teaching}, XSum~\cite{narayan2018don}, OpenWebText~\cite{Gokaslan2019OpenWebText}, and Gigaword~\cite{rush2015neural, graff2003gigaword}, using word-level tokenization, providing comprehensive insights into its performance across various text domains and summarization tasks.

    \item We perform a comparative analysis of BiSparse-AAS within a GPT-2-style model, demonstrating its ability to efficiently capture long-range dependencies while maintaining a base context window of 1024 tokens.

    \item We evaluate model performance using various evaluation metrics, including perplexity, token-level F1 score, precision, recall, ROUGE scores, and BERTScore.

    \item We demonstrate the practical feasibility and applicability of the BiSparse-AAS framework in real-world, resource-constrained scenarios by highlighting its ability to balance computational efficiency and performance on large-scale datasets.

    \item We establish a foundation to extend BiSparse-AAS to longer context windows and more complex attention patterns, which will facilitate future advancements in efficient and scalable sequence modeling.
  
\end{itemize}

\section{Proposed Mechanisms}
\label{Our_Proposed-Approaches}

State-of-the-art Transformer-based summarization models face challenges including high computational cost, quadratic complexity, and limited context preservation. To address these limitations, we propose BiSparse-AAS, a framework that improves efficiency, scalability, and contextual awareness for long-sequence processing. A central component is bilinear attention, a computationally efficient alternative that reduces parameter count and optimizes resource usage. BiSparse-AAS integrates three mechanisms: \textit{Sparse Attention}, which allocates computational focus to the most relevant tokens; \textit{Adaptive Attention Spans}, which dynamically adjust attention ranges to better preserve context; and a \textit{Hybrid Approach} that combines both techniques. These mechanisms form a cohesive framework for summarizing lengthy documents. We conclude by highlighting their unified advantages in efficiency, scalability, and real-world applicability.

\subsection{Bilinear Attention}

Bilinear attention reduces the parameter count while preserving the ability to model complex token relationships. By optimizing the self-attention mechanism, this approach improves computational efficiency and enables the model to process input with fewer resources. The standard self-attention mechanism in Transformers scales quadratically with the sequence length $N$ because it requires pairwise computations between the query and key vectors. For input queries $Q \in \mathbb{R}^{N \times d}$, $K \in \mathbb{R}^{N \times d}$, and values $V \in \mathbb{R}^{N \times d}$, the attention scores are computed using a dot product, as shown in Equation~\ref{transformer_attention}. The scores are scaled by $\frac{1}{\sqrt{d_k}}$, where $d_k$ is the dimensionality of the key vectors~\cite{vaswani2017attention}, to stabilize the gradients during training.

\begin{equation} \operatorname{Attention}(Q, K, V)=\operatorname{softmax}\left(\frac{Q K^T}{\sqrt{d_k}}\right) V
\label{transformer_attention}
\end{equation}

This approach results in $O(N^2 \times d)$ complexity, which makes it computationally expensive for long sequences~\cite{vaswani2017attention}. To address this limitation, we propose a bilinear attention mechanism that introduces a learnable weight matrix $W_a \in \mathbb{R}^{d \times d}$ to replace the standard dot product with a bilinear form, as shown in Equation~\ref{bilinear_form}.

\begin{equation} \operatorname{BilinearAttention}(Q, K, V)=\operatorname{softmax}\left(\frac{Q W_a K^T}{\sqrt{d_k}}\right) V 
\label{bilinear_form}
\end{equation}

The formulation shown in Equation~\ref{bilinear_form} maintains the structure of standard attention but introduces pairwise interactions through $W_a$. Generally, the bilinear attention mechanism can be expressed as shown in Equation~\ref{bilinear_form2} where $B(Q, K)$ represents a bilinear transformation of $Q$ and $K$. For example, if $B(Q, K)$ is defined as $Q W_a K^T$, Equation~\ref{bilinear_form} and Equation~\ref{bilinear_form2} become mathematically equivalent. Alternatively, omitting the softmax normalization step simplifies the attention mechanism, as shown in Equation~\ref{bilinear_form_alternative}.

\begin{equation}
\operatorname{BilinearAttention}(Q, K, V)=\operatorname{softmax}\left(\frac{B(Q, K)}{\sqrt{d_k}}\right) V 
\label{bilinear_form2}
\end{equation}

\begin{equation}
\operatorname{BilinearAttention}(Q, K, V)=B(Q, K) V
\label{bilinear_form_alternative}
\end{equation}

The bilinear mechanism scales linearly with $N$, reducing computational complexity to $O(N\times d)$ by projecting inputs into lower-dimensional spaces and using efficient matrix operations, enabling long-sequence modeling with lower memory and compute costs.

\subsection{Sparse Attention}

Sparse attention reduces computational overhead by focusing only on the most relevant parts of the sequence. It introduces a \emph{sparsity threshold} parameter, which determines the cutoff point for relevance. Attention scores below this threshold are masked (set to zero), reducing unnecessary computations and prioritizing tokens with higher relevance. This mechanism effectively reduces cross-token computations by focusing on a smaller set of relevant tokens, where only meaningful connections are retained. The sparsity threshold is dynamically tuned during training, with values in the range of 0.05 to 0.1 that provide stable performance across all datasets. This selective approach improves the scalability of long documents by assigning computational resources to critical relationships within the input. For summarization tasks, the sparse attention ensures that the model concentrates on the most relevant segments required to generate effective summaries. By integrating sparse attention with the bilinear attention mechanism, we optimized computational efficiency while preserving the model’s ability to capture complex relationships. The sparse attention can be formulated as shown in Equation~\ref{sparse_attention_equation}, where $M \in{0,1}^{N \times N}$ is a binary mask matrix based on the sparsity threshold. The entries of $M$ are defined in Equation~\ref{binar_mask_sparse_attention}, and $\odot$ represents element-wise multiplication. The mask $M$ selectively retains only the most relevant attention scores, which ensures computational focus on important parts of the sequence. The complexity is reduced to $O\left(s \times N \times d_k\right)$, where $s \ll N$ is the average number of elements attended per query. The scaling factor ($\frac{1}{\sqrt{d_k}}$) stabilizes the attention score computation, which is consistent with the standard transformer mechanism.

\begin{equation}
\text {Sparse}(Q, K, V)=\operatorname{softmax}\left(\frac{\left(Q W_a K^T\right) \odot M}{\sqrt{d_k}}\right) V
\label{sparse_attention_equation}
\end{equation}

\begin{equation}
M_{i, j}= \begin{cases}1 & \text { if } \frac{Q_i K_j^T}{\sqrt{d_k}} \geq \text { sparsity threshold } \\ 0 & \text { otherwise }\end{cases}
\label{binar_mask_sparse_attention}
\end{equation}

\subsection{Adaptive Attention Spans}

Adaptive attention spans dynamically adjust the attention window for each attention head based on the context, rather than relying on a fixed window size. For a given token $i$, the learned attention span $l_i$ determines the range of indices to which it can attend. This mechanism focuses on the most relevant context in long sequences, reducing computational overhead while maintaining performance. The adaptive attention mechanism learns an attention span matrix $\mathbf{A}$, where each element is processed through a sigmoid activation to derive a mask that dynamically scales attention scores. This is expressed conceptually in Equation~\ref{adaptive_spans_attention_equation}, where $W_a$ represents the bilinear transformation matrix used to compute interactions between queries and keys, and $M$ is a mask that restricts attention to a subset of adaptive span tokens. The mask $M_{i,j}$ is dynamically determined using a learned span matrix processed through sigmoid activation and can be expressed as shown in Equation~\ref{adaptive_mask}. To allow smooth transitions and eliminate hard binary cutoffs, the mask, $M_{i,j}$, in our implementation, is softened through a sigmoid-based mechanism, as shown in Equation~\ref{adaptive_span_mask}. This approach ensures that attention scores are continuously scaled, and tokens below the $span\_drop$ threshold are effectively excluded from the computation. The $span\_drop$ hyperparameter is tuned within the range of 0.05 to 0.2 to achieve stable performance across tasks and datasets. Using sigmoid-based sparsity, the model dynamically controls attention spans, achieving computational efficiency without requiring explicit regularization terms to enforce sparsity. When integrated with the bilinear attention mechanism, adaptive spans further enhance efficiency by aligning the dynamic adjustment of attention ranges with the model’s ability to capture complex token relationships. The bilinear component ($W_a$) ensures that computational resources are focused not only on task-relevant tokens, but also on their meaningful interactions, making the mechanism computationally efficient and performance-driven.

\begin{equation}
\text {Adaptive}(Q, K, V)=\operatorname{softmax}\left(\frac{\left(Q W_a K^T\right) \odot M}{\sqrt{d_k}}\right) V
\label{adaptive_spans_attention_equation}
\end{equation}

\begin{equation}
M_{i, j}= \begin{cases}1, & \text { if } j \geq \max \left(0, i-l_i\right) \text { and } j \leq i, \\ 0, & \text { otherwise. }\end{cases}
\label{adaptive_mask}
\end{equation}

\begin{equation}
M_{i, j} = \operatorname{Sigmoid}\left(\mathbf{A}_{i, j}\right) > \text{span\_drop}
\label{adaptive_span_mask}
\end{equation}

\subsection{Hybrid Approach}

The proposed hybrid approach integrates sparse attention and adaptive attention spans to leverage their complementary strengths, enhancing both computational efficiency and model performance. Sparse attention reduces computational overhead by masking irrelevant parts of the input. In contrast, adaptive spans dynamically adjust the attention window for each head based on the context, ensuring that attention spans are fine-tuned to the requirements of the task and input structure. These mechanisms are applied sequentially within the same attention heads: sparse attention first masks out less relevant tokens based on the sparsity threshold, and adaptive attention spans refine the focus further by dynamically adjusting the attention range within the remaining relevant tokens. This sequential integration ensures that computational resources are efficiently allocated to the most relevant and contextually important parts of the input. This approach is particularly advantageous for tasks that require both long-range dependencies and a fine-grained focus on critical input segments, such as document summarization and long-form text generation. The hybrid attention mechanism can be expressed as shown in Equation~\ref{hybrid_approach_equation}, where $Q W_a K^T$ represents the bilinear attention mechanism, $M_s$ denotes the sparsity mask derived from the sparse attention mechanism, and $M_a$ represents the adaptive mask dynamically adjusted using sigmoid-based spans. By capturing both global and local dependencies, the hybrid approach enables the model to process long sequences effectively without sacrificing the performance of fine-grained details. This combination is well-suited for tasks that require both scalability and precision.

\begin{equation}
\operatorname{Hybrid}(Q, K, V)=\operatorname{softmax}\left(\frac{\left(Q W_a K^T\right) \odot M_s \odot M_a}{\sqrt{d_k}}\right) V
\label{hybrid_approach_equation}
\end{equation}

\subsection{Unified Advantages of Proposed Mechanisms}
Our proposed mechanisms offer a unified framework for addressing the key challenges associated with text summarization. By integrating their complementary strengths, the proposed framework offers the following advantages.

\vspace{1.0ex}

\noindent \textbf{Computational Efficiency}. Bilinear attention reduces the parameter count and forms a lightweight base for sparse attention and adaptive spans.

\vspace{1.0ex}
\noindent \textbf{Scalability}. Sparse attention improves scalability by focusing computations on the most relevant parts of the input sequence. This approach allows the framework to handle long documents efficiently, reducing memory and computational costs.

\vspace{1.0ex}

\noindent \textbf{Context Preservation}. Adaptive attention spans dynamically adjust attention ranges for each head to ensure that critical contextual information is preserved, even for lengthy sequences. This capability improves the coherence and accuracy of generated summaries, which is a primary challenge in text summarization.

\vspace{1.0ex}

\noindent \textbf{Task-Specific Focus}. The hybrid approach integrates sparse attention and adaptive spans, balancing long-range dependency modeling with a fine-grained focus on key input segments. This makes the proposed framework particularly effective for summarization tasks that require selective emphasis on the most relevant parts of a document.

\vspace{1.0ex}

\noindent \textbf{Real-World Applicability}. The proposed framework is designed for real-world scenarios, including deployment in resource-constrained environments. Its computational efficiency and scalability make this method suitable for diverse applications, such as large-scale text summarization, where resources are often limited. The BiSparse-AAS modules are designed as drop-in replacements for Transformer self-attention layers, making them easily integrable into architectures such as BART~\cite{lewis2020bart} or T5~\cite{raffel2020exploring} with minimal modification to encoder or decoder blocks.

\section{Model Architecture}
\label{model_architecture}

\begin{figure}[htbp]
  \centering
  \includegraphics[width=0.8\columnwidth]{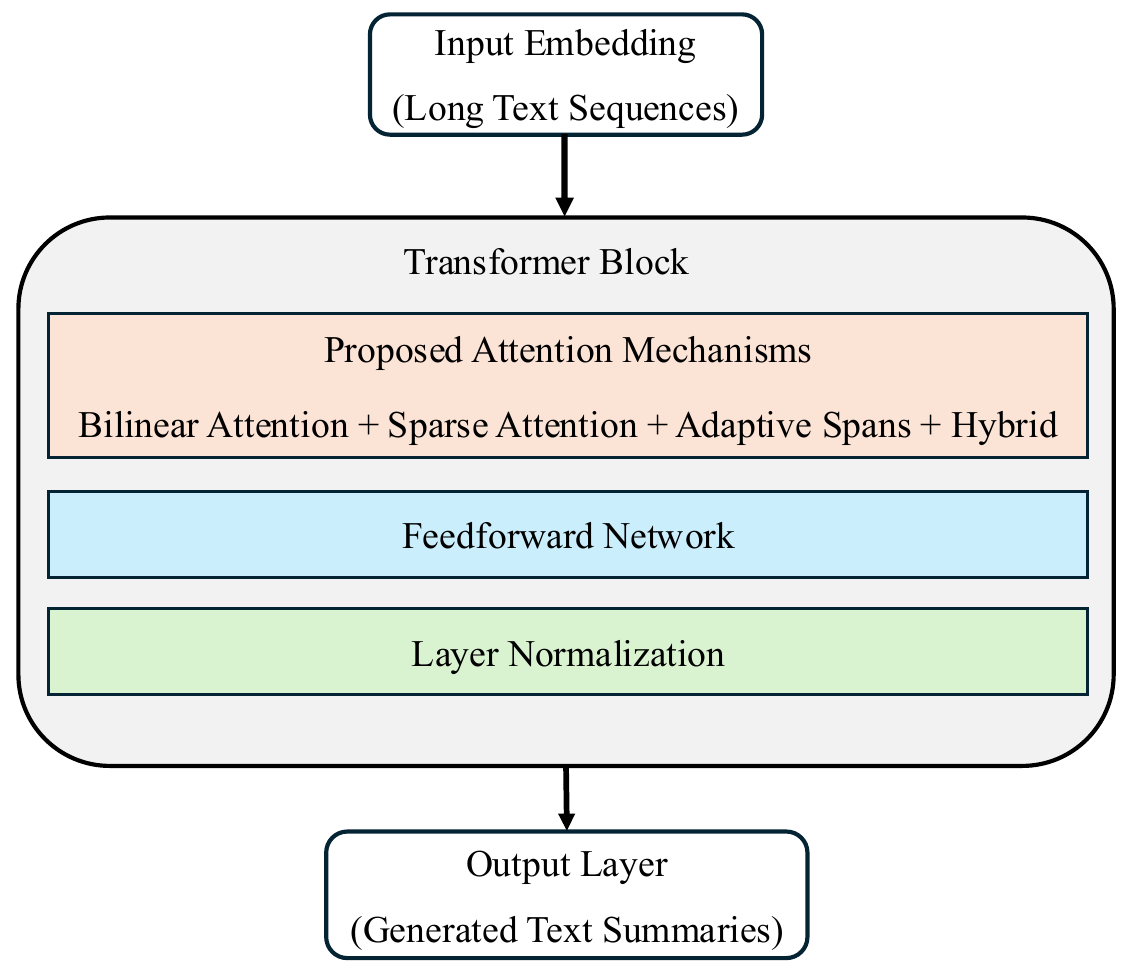}
  \caption{Model architecture.}
  \label{Fig:model_architecture}
  \vspace{-1.5ex}
\end{figure}

The BiSparse-AAS architecture, shown in Fig.~\ref{Fig:model_architecture}, is designed to address key challenges in text summarization, particularly in the handling of long sequences, improving computational efficiency, and preserving contextual integrity. The model consists of the following components:

\subsection{Input Processing}

The input sequence is preprocessed to ensure consistency, stability, and efficiency during the learning process. The input tokens are first mapped to their respective embeddings, and positional encodings are added to incorporate semantic and sequential information. Layer normalization is applied to the combined input embeddings to stabilize leading dynamics, facilitate efficient learning, and mitigate the risk of exploding or vanishing gradients. The normalized embeddings are then passed to the attention mechanisms, where the model computes relevance scores to identify and selectively focus on the most critical parts of the input sequence. This preprocessing ensures that the model effectively handles long sequences while preserving essential contextual information.

\subsection{Transformer Block}

The Transformer block (see Figure~\ref{Fig:model_architecture}) serves as the basis of the BiSparse-AAS model, capturing complex relationships within input sequences. It retains standard residual connections around the attention and feedforward sub-layers to ensure gradient flow and stable training, with no deviations from the core Transformer design. The block comprises the following components:

\vspace{1.0ex}
\noindent \textbf{Attention Mechanisms}. The attention layer integrates bilinear attention, sparse attention, adaptive attention spans, and their combination in the hybrid approach. Using these attention mechanisms, the model achieves efficient relevance computation, scalability, and robust context preservation across the input. Each block employs a single attention mechanism, which is determined by the model configuration. This mechanism allows flexibility in switching between sparse, adaptive, or hybrid attention modes at the block level. This design ensures modularity without introducing overlapping mechanisms within the same block. Each mechanism addresses specific challenges in text summarization tasks:

\begin{itemize}
    \item \textbf{Bilinear Attention}. Reduces parameter count and improves computational efficiency while preserving the ability to capture complex token relationships.

    \item \textbf{Sparse Attention}. By reducing computational demands and focusing on the most relevant parts of the input tokens, sparse attention minimizes unnecessary overhead. This enables the model to scale efficiently to handle long sequence inputs.

    \item \textbf{Adaptive Attention Spans}. Adaptive spans dynamically adjust attention ranges for each attention head to handle long sequences without excessive memory consumption. This ensures that critical contextual information is preserved even in long sequences.

   \item \textbf{Hybrid Approach}. Combines sparse attention and adaptive spans to balance long-range dependencies with fine-grained focus on critical input segments, ensuring efficiency and accuracy.

\end{itemize}

\vspace{1.0ex} 
\noindent \textbf{Positional Encoding}. The model uses standard learned positional embeddings added to token embeddings, ensuring compatibility with GPT architecture and being expressive enough for text summarization tasks.

\vspace{1.0ex}
\noindent \textbf{Feedforward Network}. A fully connected feedforward network processes each token position independently through a series of dense layers, introducing nonlinearity and enabling the model to learn complex feature transformations from the input. It complements the attention mechanisms by refining token-level representations critical for capturing hierarchical and semantic relationships in the input sequence.

\vspace{1.0ex}
\noindent \textbf{Layer Normalization}. Applied within the Transformer block, layer normalization stabilizes training and accelerates convergence by reducing internal covariate shifts. It improves robustness and adaptability, enabling the model to generalize well to diverse datasets and unseen sequence lengths.

\subsection{Output Processing}

The output of the Transformer block is passed through a final fully connected projection layer that translates the aggregated sequence representation into the desired output format. This layer ensures that the generated summaries are coherent, contextually relevant, and concise, leveraging the model’s ability to balance long-range context with detailed focus to preserve key information from the input sequence.

\vspace{1.0ex}
\noindent \textbf{Coherence}. The output processing ensures that the generated summaries maintain logical flow and grammatical correctness.

\vspace{1.0ex}
\noindent \textbf{Task-Specific Relevance}. Attention scores and intermediate representations are fine-tuned to focus on task-specific objectives, such as extractive or abstractive summarization.

\section{Experimental Settings}
\label{experimental_settings}

This section describes the datasets and the implementation and evaluation strategies used in our experiments.

\subsection{Datasets}

We evaluated our proposed BiSparse-AAS model on four benchmark datasets: XSum~\cite{narayan2018don}, CNN/DailyMail~\cite{hermann2015teaching}, OpenWebText~\cite{Gokaslan2019OpenWebText}, and Gigaword~\cite{rush2015neural, graff2003gigaword}.

\vspace{1.0ex}
\noindent \textbf{CNN/DailyMail}. This dataset contains approximately 300K news articles from CNN and the Daily Mail, each paired with multi-sentence highlights~\cite{hermann2015teaching}. The summaries are largely extractive, often consisting of sentences extracted directly from the source text, making it well-suited for evaluating extractive models. While also applicable to abstractive summarization, models trained on this dataset should aim to generate novel and coherent summaries rather than over-relying on copying source sentences.

\vspace{1.0ex}
\noindent \textbf{XSum}. XSum comprises approximately 227K BBC news articles, each paired with a single-sentence summary~\cite{narayan2018don}. The summaries are highly abstractive, requiring the model to rephrase, generalize, and compress the input text. This dataset is particularly challenging, as it demands deep semantic understanding and advanced text generation capabilities.

\vspace{1.0ex}
\noindent \textbf{OpenWebText}. The OpenWebText dataset~\cite{Gokaslan2019OpenWebText}, originally created for general language modeling tasks, lacks predefined summaries and is not directly suited for standard summarization tasks. Since OpenWebText does not provide predefined summaries, we followed a common adaptation by splitting each document into two halves: the first half was treated as the ``article'' and the second as the ``summary''. Although this heuristic is subjective, it enables consistent large-scale evaluation and has precedent in prior summarization adaptations. To preprocess the text, we applied a word-level GPT-2 tokenizer. This setup simulates an abstractive summarization task, enabling meaningful evaluation in this context. Given its diverse content and varied text lengths, OpenWebText serves as a valuable testbed for assessing short- and long-range dependencies. This diversity aligns well with the strengths of our proposed mechanisms, which perform consistently across all evaluation metrics, as shown in Table~\ref{tab:metrics-comparison}.

\vspace{1.0ex}
\noindent \textbf{Gigaword}. The Gigaword dataset, derived from over 4.3 million news articles in the English Gigaword corpus provided by the Linguistic Data Consortium, contains millions of sentence pairs, each consisting of a news sentence and its corresponding headline~\cite{graff2003gigaword}. These short and concise pairs make Gigaword suitable for evaluating models focused on abstractive summarization and headline generation. Unlike datasets with multi-sentence summaries, it emphasizes single-sentence condensation, requiring models to generate compact summaries that capture the core meaning of the input.

\subsection{Data Handling}

\noindent \textbf{Tokenization and Batching}. We used a word-level GPT-2 tokenizer with a fixed vocabulary size of 50,257 across all datasets. The documents were tokenized into subword units, and padding was applied to standardize the sequence lengths within each batch. To optimize GPU memory usage and minimize padding overhead, examples with similar lengths were grouped into the same batch.

\vspace{1.0ex}

\noindent \textbf{Sequence Lengths}. For XSum, the input sequences were limited to 512 tokens, with longer documents truncated as needed. This limit aligns with typical constraints of many Transformer-based models (\eg BERT~\cite{devlin2018bert}), balances computational cost, and reflects that the most salient information in XSum usually appears early in the text. For CNN/DailyMail, OpenWebText, and Gigaword, input sequences were set to a maximum of 1024 tokens to accommodate longer documents while maintaining efficiency within our model's base context window. During training, the number of tokens processed per iteration was 294,912 for XSum and 589,824 for the other three datasets, consistent with their respective sequence lengths. This setup demonstrates our model’s ability to efficiently handle both short and long input sequences.

\vspace{1.0ex}

\noindent \textbf{Dataset Splits}. We followed the standard training, validation, and test splits for each dataset to ensure fair comparisons and reproducibility of the results.

\subsection{Model Configuration and Training}

\noindent \textbf{Model Parameters}. The BiSparse-AAS model was configured with 12 attention heads and a base context window of 1024 tokens. The sparsity and span drop thresholds were dynamically tuned in the validation sets to determine optimal values for each dataset. We found that a sparsity threshold of 0.1 and a span drop threshold of 0.2 consistently produced the best performance across all datasets. These settings strike a balance between computational efficiency and context preservation, consistent with the findings of previous studies on sparse attention and adaptive spans~\cite{treviso2022predicting, sukhbaatar2019adaptive}. This hybrid approach ensures that the model remains adaptable to varying dataset characteristics. The parameter count remained nearly consistent across all configurations: sparse attention used approximately 102.91 million parameters, while adaptive spans and the hybrid mechanism required about 103.06 million (see Table~\ref{table:parameter_count-tokens-per-iteration}). The slightly lower parameter count for sparse attention is due to its masking mechanism, which reduces the need for additional learned weights. Hybrid and adaptive configurations introduced only a marginal increase of $\sim$0.15 million parameters, demonstrating their ability to improve performance with minimal computational overhead.

\vspace{1.0ex}
\noindent \textbf{Optimization}. The model was trained using Adam optimizer with a learning rate of $6e^{-4}$ and a linear decay schedule. Early stopping based on validation loss was applied with a patience value of 5. To simulate a larger effective batch size, we used a gradient accumulation step size of 48 ($6 \times 8$). The training was run for up to 500,000 iterations, with a fixed batch size of 12 to support larger input sequences while optimizing GPU memory usage.

\vspace{1.0ex}

\noindent \textbf{Hardware Setup}. All experiments were carried out on two NVIDIA GeForce RTX 4090 GPUs, each with 24 GB of GDDR6X memory. The training process was parallelized using PyTorch’s distributed data-parallel (DDP) framework to enable efficient use of multiple GPUs.

\subsection{Evaluation Metrics}

We evaluated the model using several standard metrics to assess the coherence, precision, and semantic richness of the generated summaries.

\vspace{1.0ex}
\noindent \textbf{Perplexity}. Measures the model’s ability to predict the next token in a sequence, where lower values indicate better predictive performance and model confidence.

\vspace{1.0ex}
\noindent \textbf{ROUGE-1/2/L}. Evaluates the lexical overlap between the generated and reference summaries using unigrams, bigrams, and longest common subsequences.

\vspace{1.0ex}
\noindent \textbf{BERTScore}. Computes the semantic similarity between generated and reference summaries by comparing their contextual embeddings from a pre-trained BERT model.

\vspace{1.0ex}
\noindent \textbf{Token-Level Precision, Recall, F1}. Measures the accuracy of generating relevant tokens by evaluating how many were correctly produced (precision), how many relevant tokens were retrieved (recall), and their harmonic mean (F1-score).

\begin{figure*}[t]
  \centering
  \begin{subfigure}{0.80\linewidth}
    \includegraphics[width=\linewidth]{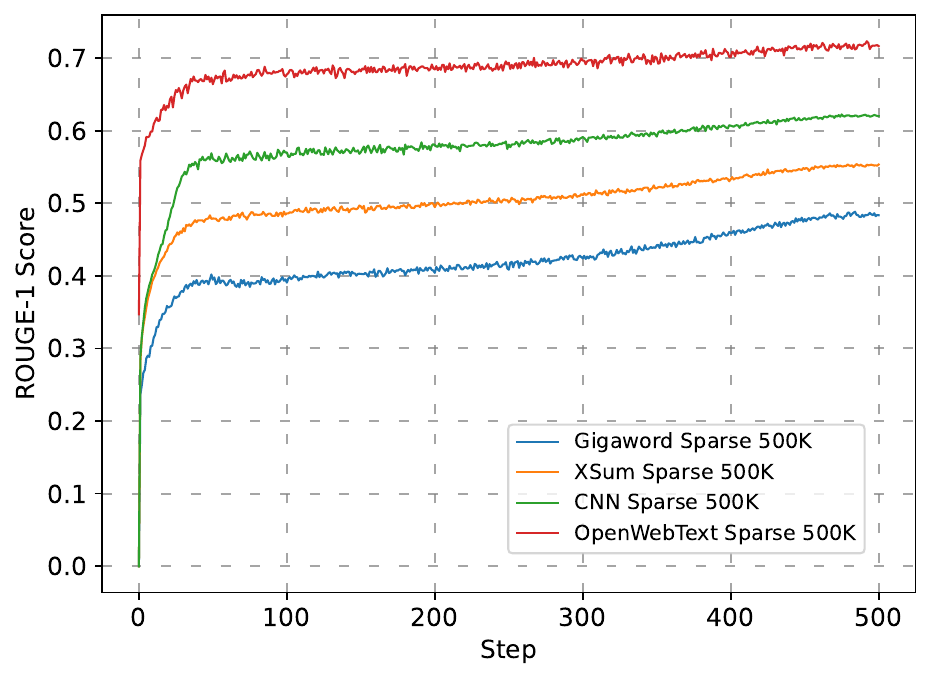}
    \caption{Sparse Attention: ROUGE-1 Scores}\label{fig:rouge1sparse}
  \end{subfigure}\hfill
  \begin{subfigure}{0.80\linewidth}
    \includegraphics[width=\linewidth]{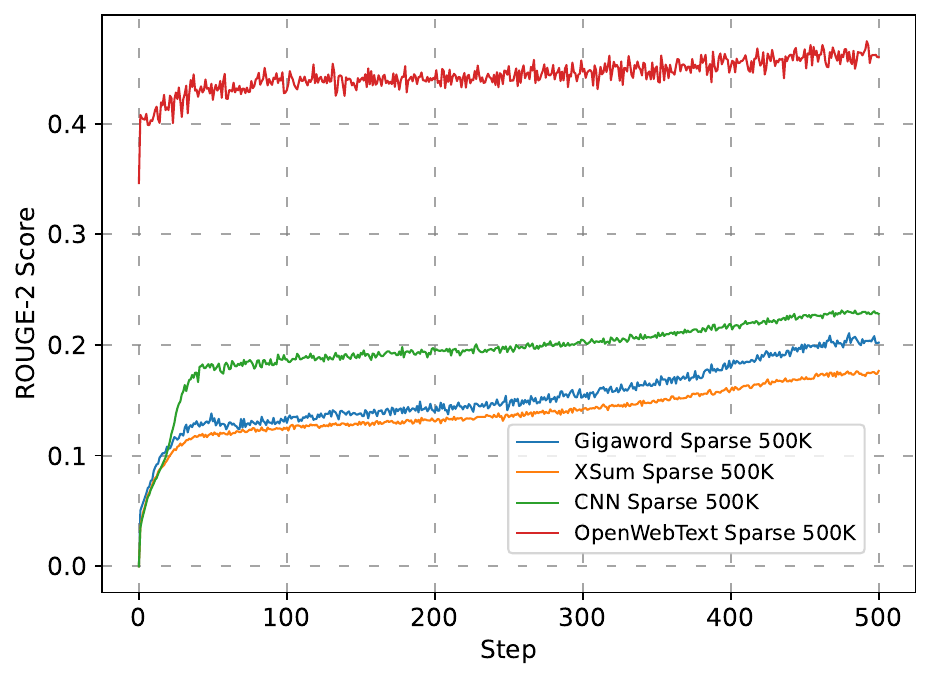}
    \caption{Sparse Attention: ROUGE-2 Scores}\label{fig:rouge2sparse}
  \end{subfigure}
 
  \label{fig:rouge-values}
\end{figure*}

\begin{figure*}[t]
  \ContinuedFloat
  \centering
  \begin{subfigure}{0.80\linewidth}
    \includegraphics[width=\linewidth]{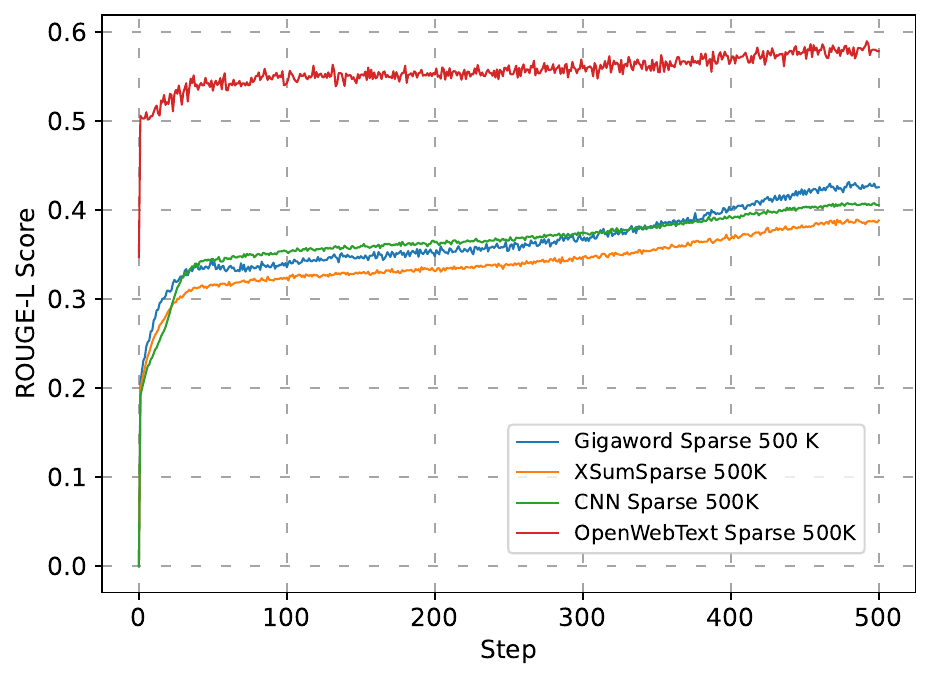}
    \caption{Sparse Attention: ROUGE-L Scores}\label{fig:rougeLsparse}
  \end{subfigure}\hfill
  \begin{subfigure}{0.80\linewidth}
    \includegraphics[width=\linewidth]{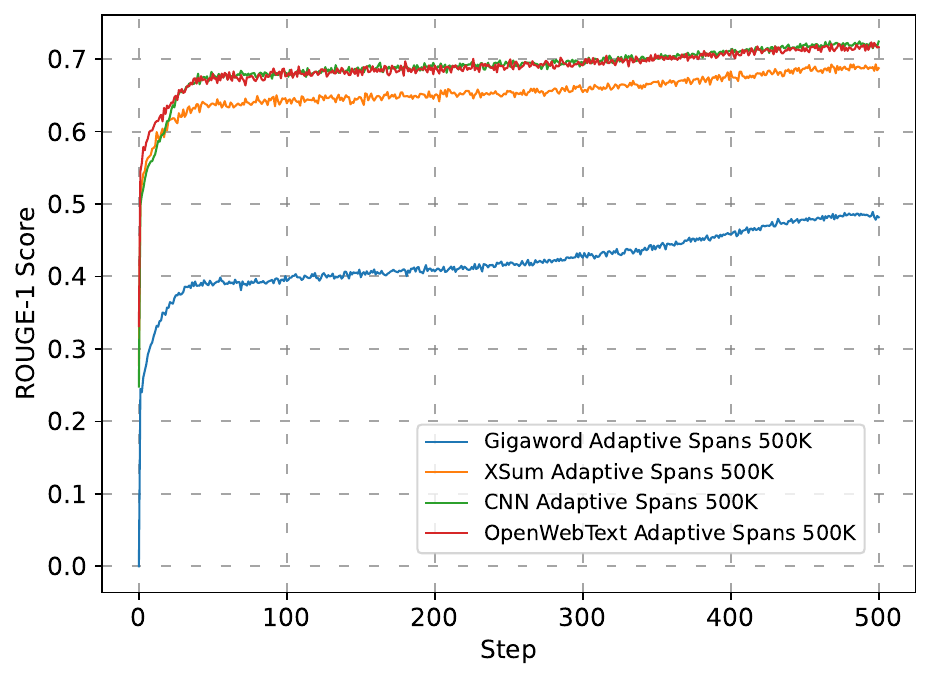}
    \caption{Adaptive Spans: ROUGE-1 Scores}\label{fig:rouge1adaptive}
  \end{subfigure}
  \caption{Validation ROUGE scores across 500K training steps for sparse attention and adaptive spans. The curves illustrate the training dynamics and stability of each mechanism, complementing the dataset-specific results discussed in Section~\ref{results_and_discussion}. Due to space limitations, we report R-1, R-2, and R-L scores for sparse attention and the R-1 score for adaptive spans.}

\label{fig:rouge-values}
\end{figure*}

%%% TABLES %%%%

\begin{table*}[htbp]
\centering
\setlength{\tabcolsep}{3pt}  
\caption{Metrics for Sparse Attention, Adaptive Spans, and Hybrid approaches on CNN/DailyMail, XSum, OpenWebText, and Gigaword Datasets. \textbf{Note}: F1-score, Precision, and Recall are Token-Level metrics. ``Hybrid'' refers to the sequential integration of sparse attention and adaptive spans within the same attention heads.}
\label{tab:metrics-comparison}
\resizebox{\textwidth}{!}{
\begin{tabular}{llccccccccc}

\toprule
\textbf{Dataset} & \textbf{Approach} & \textbf{Perplexity} & \textbf{F1-score} & \textbf{Precision} & \textbf{Recall} & \textbf{R1} & \textbf{R2} & \textbf{RL} & \textbf{BERTScore} \\

\midrule
\multirow{3}{*}{CNN/DailyMail~\cite{hermann2015teaching}} 
 & Sparse Attention & 9.70 & 54.60 & 56.00 & 56.35 & 62.00 & 22.80 & 40.53 & 83.10 \\
 & Adaptive Spans & 9.37 & 55.32 & 56.74 & 57.05 & 72.42 & 43.63 & 56.39 & 83.58 \\
 & Hybrid Approach & 9.85 & 54.45 & 55.91 & 56.20 & 71.84 & 42.52 & 55.42 & 83.52 \\

\midrule

\multirow{3}{*}{XSum~\cite{narayan2018don}} & Sparse Attention & 8.98 & 55.60 & 55.98 & 57.88 & 55.34 & 17.65 & 38.80 & 83.73 \\
 & Adaptive Spans & 9.00 & 55.64 & 56.00 & 57.95 & 68.74 & 41.51 & 56.21 & 87.03 \\
 & Hybrid Approach & 8.90 & 55.77 & 56.14 & 58.05 & 68.85 & 41.64 & 56.30 & 87.07 \\

\midrule
\multirow{3}{*}{OpenWebText~\cite{Gokaslan2019OpenWebText}} 
 & Sparse Attention & 9.92 & 56.32 & 57.11 & 58.68 & 71.65 & 46.00 & 57.80 & 83.24 \\
 & Adaptive Spans & 9.91 & 56.37 & 57.26 & 58.71 & 71.66 & 46.07 & 57.85 & 83.17 \\
 & Hybrid Approach & 9.88 & 56.62 & 57.51 & 58.90 & 71.78 & 46.36 & 58.08 & 83.30 \\

\midrule
\multirow{3}{*}{Gigaword~\cite{rush2015neural, graff2003gigaword}} 
 & Sparse Attention & 1.16 & 97.29 & 97.31 & 97.45 & 48.35 & 20.21 & 42.55 & 87.69 \\
 & Adaptive Spans & 1.15 & 97.30 & 97.31 & 97.46 & 48.17 & 19.88 & 42.34 & 87.73 \\
 & Hybrid Approach & 1.15 & 97.30 & 97.31 & 97.46 & 48.87 & 20.68 & 42.96 & 87.79 \\

\bottomrule
\end{tabular}%
}

\end{table*}

\begin{table*}[htbp]
\centering

\caption{Tokens processed per iteration and parameter counts for BiSparse-AAS across different datasets and attention mechanisms. Sequence length denotes the maximum number of tokens per dataset, and tokens per iteration vary based on both sequence length and dataset-specific batch sizes. ``Hybrid'' refers to the sequential integration of sparse attention and adaptive spans within the same attention heads.}

\label{table:parameter_count-tokens-per-iteration}
\resizebox{\textwidth}{!}{
\begin{tabular}{llccc}
\toprule
\textbf{Dataset} & \textbf{Approach} & \textbf{Sequence Length} & \textbf{Tokens Per Iteration} & \textbf{Parameters (M)} \\
\midrule
\multirow{3}{*}{XSum~\cite{narayan2018don}} 
 & Sparse Attention & 512 & 294,912 & 102.91  \\
 & Adaptive Spans & 512 & 294,912 & 102.98  \\
 & Hybrid Approach & 512 & 294,912 & 102.98 \\
\midrule
\multirow{3}{*}{CNN/DailyMail~\cite{hermann2015teaching}} 
 & Sparse Attention & 1024 & 589,824 & 102.91 \\
 & Adaptive Spans & 1024 & 589,824 & 103.06 \\
 & Hybrid Approach & 1024 & 589,824 & 103.06 \\
\midrule
\multirow{3}{*}{OpenWebText~\cite{Gokaslan2019OpenWebText}} 
 & Sparse Attention & 1024 & 589,824 & 102.91  \\
 & Adaptive Spans & 1024 & 589,824 & 103.06 \\
 & Hybrid Approach & 1024 & 589,824 & 103.06  \\

 \midrule
\multirow{3}{*}{Gigaword~\cite{rush2015neural, graff2003gigaword}} 
 & Sparse Attention & 1024 & 589,824 & 102.91  \\
 & Adaptive Spans & 1024 & 589,824 & 103.06 \\
 & Hybrid Approach & 1024 & 589,824 & 103.06  \\
 
\bottomrule
\end{tabular}
}

\end{table*}

\begin{table*}[htbp]
\centering
\setlength{\tabcolsep}{3pt}  

\caption{Comparison of BiSparse-AAS approaches with state-of-the-art baseline models on the XSum, CNN/DailyMail, OpenWebText, and Gigaword datasets using ROUGE metrics (R1, R2, RL). ``Hybrid'' refers to the sequential integration of sparse attention and adaptive spans within the same attention heads. Baselines are not reported on OpenWebText because prior models did not evaluate on this dataset.}
\label{table:comparison_bisparse_aas}
\resizebox{\textwidth}{!}{%
\begin{tabular}{l|ccc|ccc|ccc|ccc}
\toprule
\textbf{Model} & \multicolumn{3}{c|}{\textbf{XSum}} & \multicolumn{3}{c|}{\textbf{CNN/DailyMail}} & \multicolumn{3}{c|}{\textbf{OpenWebText}} & \multicolumn{3}{c}{\textbf{Gigaword}} \\
 & \textbf{R1} & \textbf{R2} & \textbf{RL} & \textbf{R1} & \textbf{R2} & \textbf{RL} & \textbf{R1} & \textbf{R2} & \textbf{RL} & \textbf{R1} & \textbf{R2} & \textbf{RL} \\
\midrule

BERTShare~\cite{rothe2020leveraging}& 38.52 & 16.12 & 31.13 & 39.25 & 18.09 & 36.45 & -- & -- & -- & 38.13 & 19.81 & 35.62 \\

BERTSUMABS~\cite{liu2019text} & 38.76 & 16.33 & 31.15 & 41.72 & 19.39 & 38.76 & -- & -- & -- & -- & -- & -- \\
BERTSUMEXTABS~\cite{liu2019text} & 38.81 & 16.50 & 31.27 & 41.13 & 19.60 & 39.18 & -- & -- & -- & -- & -- & -- \\
  
BART~\cite{lewis2020bart} & 45.14 & 22.27 & 37.25 & 44.16 & 21.28 & 40.90 & -- & -- & -- & -- & -- & -- \\
T5~\cite{raffel2020exploring} & -- & -- & -- & 43.52 & 21.55 & 40.69 & -- & -- & -- & -- & -- & --\\
LEAD~\cite{zhu2020make} & 16.30 & 1.60 & 11.95 & 40.34 & 17.70 & 36.57 & -- & -- & -- & 21.86 & 7.66 & 20.45\\
LEAD-3~\cite{zhu2020make, dong2019unified} & 16.30 & 1.60 & 11.95 & 40.42 & 17.62 & 36.67 & -- & -- & -- & 21.86 & 7.66 & 20.45\\

UNILM~\cite{dong2019unified} & -- & -- & -- & 43.33 & 20.21 & 40.51 & -- & -- & -- & 38.45 & 19.45 & 35.75\\
ProphetNet~\cite{qietal2020ProphetNet} & -- & -- & -- & 43.68 & 20.64 & 40.72 & -- & -- & -- & 39.55 & 20.27 & 36.57\\
PTGen~\cite{see2017get} & 29.70 & 9.21 & 23.24 & 36.44 & 15.66 & 33.42 & -- & -- & -- & -- & -- & --\\ 
PTGen+Coverage~\cite{see2017get, narayan2018don} & 28.10 & 8.02 & 21.72 & 39.53 & 17.28 & 36.38 & -- & -- & -- & -- & -- & --\\ 

MASS~\cite{song2019mass} & 39.75 & 17.24 & 31.95 & 42.12 & 19.50& 39.01 & -- & -- & -- & 38.73 & 19.71 & 35.96\\

PEGASUS\textsubscript{LARGE} (C4)~\cite{zhang2020pegasus} & 45.20 & 22.06 & 36.99 & 43.90 & 21.20 & 40.76 & -- & -- & -- & 38.75 & 19.96 & 36.14\\

PEGASUS\textsubscript{LARGE} (HugeNews)~\cite{zhang2020pegasus} & 47.21 & 24.56 & 39.25 & 44.17 & 21.47 & 41.11 & -- & -- & -- & 39.12 & 19.86 & 36.24\\

\midrule

\multicolumn{13}{l}{\textbf{BiSparse-AAS}} \\
Sparse Attention & 55.34 & 17.65 & 38.80 & 62.00 & 22.80 & 40.53 & 71.65 & 46.00 & 57.80 & 48.35 & 20.21 & 42.55 \\
Adaptive Spans   & 68.74 & 41.51 & 56.21 & 72.42 & 43.63 & 56.39 & 71.66 & 46.07 & 57.85 & 48.17 & 19.88 & 42.34 \\
Hybrid Approach  & 68.85 & 41.64 & 56.30 & 71.84 & 42.52 & 55.42 & 71.78 & 46.36 & 58.08 & 48.87 & 20.68 & 42.96 \\
\bottomrule
\end{tabular}
}

\end{table*}

\section{Results and Discussion}
\label{results_and_discussion}

This section presents a comprehensive evaluation of BiSparse-AAS across multiple benchmark datasets, comparing its performance with state-of-the-art baselines to assess both efficiency and effectiveness.

\subsection{Performance Across Datasets}

The experimental results presented in Tables~\ref{tab:metrics-comparison} and~\ref{table:comparison_bisparse_aas}, along with the ROUGE plots shown in Fig.~\ref{fig:rouge-values}, demonstrate consistent performance improvements across all datasets. For clarity, the ``Hybrid'' model reported in our tables and figures refers to a sequential integration of sparse attention and adaptive spans within the same attention heads.

\vspace{1.0ex} 
\noindent \textbf{CNN/DailyMail}. Adaptive spans achieved the highest performance (R1: 72.42, R2: 43.63, RL: 56.39), significantly outperforming both sparse attention and the hybrid approach. The hybrid method yielded comparable results, with slightly lower scores (R2: 42.52, RL: 55.42), while sparse attention showed the lowest results (R1: 62.00, R2: 22.80, RL: 40.53). These findings show the advantage of dynamic span adjustment in preserving contextual information for extractive summarization of long sequences. The associated BERTScore demonstrates its ability to generate semantically coherent summaries. Figure~\ref{fig:rouge-values} provides a complementary view of model convergence during training, showing stable improvements for adaptive spans on CNN/DailyMail.

\vspace{1.0ex}
\noindent \textbf{XSum}. For abstractive single-sentence summarization, the hybrid approach performed best (see Table~\ref{tab:metrics-comparison}), with R1, R2, and RL values of 68.85, 41.64, and 56.30, respectively. Adaptive spans followed closely with R1: 68.74, R2: 41.51, and RL: 56.21. This demonstrates the ability of the hybrid model to balance long-range dependencies with a fine-grained focus, which is critical for high-quality abstractive summarization. Sparse attention, while efficient, underperformed in this setting, scoring R1: 55.34, R2: 17.65, and RL: 38.80.

\vspace{1.0ex}
\noindent \textbf{OpenWebText}. All approaches performed comparably, with the hybrid model slightly outperforming the others (R1: 71.78, R2: 46.36, and RL: 58.08). The consistently high token-level precision, recall, and F1 scores across models demonstrate the robustness of BiSparse-AAS on diverse and large-scale text. Given OpenWebText’s broad coverage of text types, the hybrid approach benefits from its ability to balance short-range and long-range dependencies, capturing both local details and global context, which explains its advantage in this broad-coverage dataset, as reflected in Fig.~\ref{fig:rouge-values}.

\vspace{1.0ex}
\noindent \textbf{Gigaword}. As a simpler headline-generation task, all three approaches achieved high token-level metrics, with token-level precision, recall, and F1 exceeding 97\%. Sparse attention reached a perplexity of 1.16, while adaptive spans and the hybrid slightly improved it to 1.15. The hybrid also achieved higher ROUGE scores (R1: 48.87, R2: 20.68, RL: 42.96), demonstrating the efficiency and precision of BiSparse-AAS in compact text generation (see the validation results in Fig.~\ref{fig:rouge-values}).

Our experimental results suggest that adaptive spans are particularly effective for extractive summarization tasks such as CNN/DailyMail, while the hybrid mechanism excels in abstractive summarization settings like XSum that require concise outputs. Sparse attention, although less competitive in abstractive summarization, remains useful when efficiency and scalability are prioritized. These findings indicate that the choice of mechanism can be tuned based on the dataset and task requirements.

\subsection{Comparative Analysis with Baseline Models}

Across CNN/DailyMail and XSum, BiSparse-AAS substantially outperformed strong baselines such as PEGASUS~\cite{zhang2020pegasus}, BERTSUMABS~\cite{liu2019text}, and BART~\cite{lewis2020bart}. In CNN/DailyMail, the adaptive spans approach achieved R1: 72.42, R2: 43.63, RL: 56.39, compared to PEGASUS (R1: 44.17, R2: 21.47, RL: 41.11) and BART (R1: 44.16, R2: 21.28, RL: 40.90). In the XSum dataset, the hybrid approach reached R1: 68.85, R2: 41.64, RL: 56.30, significantly outperforming PEGASUS (R1: 47.21, R2: 24.56, RL: 39.25) and BART (R1: 45.14, R2: 22.27, RL: 37.25). Because most prior baselines only report ROUGE scores, so we restrict comparisons to ROUGE for fairness. Our improvements result from combining bilinear attention with sparse and adaptive mechanisms that preserve context in long documents more effectively than standard Transformer models while maintaining computational efficiency.

\subsection{Efficiency and Scalability}

Bilinear attention reduced the parameter count to 102.91M, compared to 124M in GPT-2. Adaptive spans and the hybrid increased this slightly to 103.06M due to learnable span masks. These masks allow each head to adjust its receptive field, preserving context in long sequences while maintaining efficiency. BiSparse-AAS processed up to 589,824 tokens per iteration on CNN/DailyMail, OpenWebText, and Gigaword, and 294,912 tokens on XSum, demonstrating scalability across diverse sequence lengths. In terms of runtime, BiSparse-AAS introduces negligible overhead compared to standard attention, with inference latency remaining comparable to the baseline on 2×RTX 4090 GPUs. Training throughput reached up to 589k tokens per iteration, confirming that the framework maintains both scalability and practical deployability. Although competitor implementations did not report comparable efficiency metrics, BiSparse-AAS reduced the parameter count by approximately 17\% relative to GPT-2. This efficiency makes it well-suited for deployment in resource-constrained environments.

\noindent Although we primarily report quantitative metrics, future work will include more qualitative analyses of generated summaries to assess coherence and fluency. In addition, future work will apply statistical significance tests across multiple runs to provide additional rigor in comparing models.

\section{Conclusions and Future Work}
\label{conclusion}

We introduced BiSparse-AAS, a novel framework that integrates bilinear attention, sparse attention, and adaptive attention spans for efficient text summarization. By addressing the challenges of long-sequence processing and context preservation, BiSparse-AAS provides a unified and scalable solution for both extractive and abstractive summarization tasks. Evaluations on XSum, CNN/DailyMail, OpenWebText, and Gigaword show that it achieves strong summarization performance with lower computational cost, making it well-suited for deployment in resource-constrained settings. Beyond text summarization, BiSparse-AAS offers a general framework for scalable sequence modeling, with potential applications in machine translation, long-form generation, and other NLP tasks where preserving context is critical. Despite integrating multiple mechanisms, the framework remains modular and interpretable, as each component can be enabled or disabled independently. 

Although our framework shows strong results within a 1024-token context, extending it to very long legal or scientific documents and to low-resource domains remains future work. We also plan more granular ablation studies, such as comparing bilinear and standard attention and varying span/sparsity thresholds, to better isolate individual contributions and identify opportunities for further optimization and complexity reduction. Further directions include extending the context window, optimizing sparsity thresholds, and exploring hardware-specific adaptations. Finally, by lowering computational requirements, BiSparse-AAS can help reduce the environmental impact of large-scale NLP systems, while fairness-aware training and bias mitigation could support more reliable and ethical use.

\bibliography{TextSummarization}

\bibliographystyle{plain} 

\end{document}